# A Low-cost and Ultra-lightweight Binary Neural Network for Traffic Signal Recognition

Mingke Xiao
School of Intelligent Science and Technology, Hangzhou Institute for Advanced Study, University of Chinese Academy of Sciences
Hangzhou, China
xiaomingke23@mails.ucas.ac.cn

Yue Su
School of Intelligent Science and Technology, Hangzhou Institute for Advanced Study, University of Chinese Academy of Sciences
Hangzhou, China
yuesu@ucas.ac.cn

Liang Yu
School of Intelligent Science and Technology, Hangzhou Institute for Advanced Study, University of Chinese Academy of Sciences
Hangzhou, China
yuliang23@mails.ucas.ac.cn

Guanglong Qu
School of Integrated Circuits
Southeast University
*(of Affiliation)*
Nanjing, China
220236503@seu.edu.cn

Yutong Jia
College of electronic science & engineering
Jiling University
Jiling, China
jiayt1922@mails.jlu.edu.cn

Yukuan Chang
School of Intelligent Science and Technology, Hangzhou Institute for Advanced Study, University of Chinese Academy of Sciences
Hangzhou, China
changyukuan@ucas.ac.cn

Xu Zhang
Institute of Semiconductors, Chinese Academy of Sciences
Beijing, China
zhangxu@semi.ac.cn

***Abstract*—The deployment of neural networks in vehicle platforms and wearable Artificial Intelligence-of-Things (AIOT) scenarios has become a research area that has attracted much attention. With the continuous evolution of deep learning technology, many image classification models are committed to improving recognition accuracy, but this is often accompanied by problems such as large model resource usage, complex structure, and high power consumption, which makes it challenging to deploy on resource-constrained platforms. Herein, we propose an ultra-lightweight binary neural network (BNN) model designed for hardware deployment, and conduct image classification research based on the German Traffic Sign Recognition Benchmark (GTSRB) dataset. In addition, we also verify it on the Chinese Traffic Sign (CTS) and Belgian Traffic Sign (BTS) datasets. The proposed model shows excellent recognition performance with an accuracy of up to 97.64%, making it one of the best performing BNN models in the GTSRB dataset. Compared with the full-precision model, the accuracy loss is controlled within 1%, and the parameter storage overhead of the model is only 10% of that of the full-precision model. More importantly, our network model only relies on logical operations and low-bit width fixed-point addition and subtraction operations during the inference phase, which greatly simplifies the design complexity of the processing element (PE). Our research shows the great potential of BNN in the hardware deployment of computer vision models, especially in the field of computer vision tasks related to autonomous driving.**

## I. INTRODUCTION

In recent years, autonomous driving technology has made significant progress, gradually evolving from early assisted driving systems to highly or even fully automated stages. In this process, traffic sign recognition has become a core component of the autonomous driving system [1]. Compared with conventional image recognition tasks, traffic signal recognition algorithms usually need to run on a vehicle-mounted platform, which means that while pursuing high recognition accuracy and fast response, special attention must be paid to the additional resource consumption generated during the execution of the algorithm [2]. Given that cars are resource-constrained embedded scenarios [3], traffic sign recognition systems tend to use lightweight models to ensure that they can be effectively deployed and run efficiently on vehicle platforms [4].

In order to make the model lightweight, many research methods have emerged. Among them, the model pruning technology proposed in [5] has become the core principle of the API widely used in frameworks such as TensorFlow. This technology effectively reduces the complexity of calculation by setting some parameters in the model to zero, because the calculation operations related to zero can be directly ignored, thereby improving the calculation efficiency. In addition, the int8 quantization model introduced in [6] has been widely used in the industry. The quantized model can be seamlessly deployed on a variety of devices such as Android and Apple, significantly improving the deployment flexibility and operating efficiency of the model. In order to further pursue higher quantization efficiency and performance, [7] pioneered the concept of binary neural network (BNN). BNN greatly simplifies the calculation process by limiting the weights and input values in the model to {1, -1}, opening up a new way to lightweight neural network models.

In this paper, we use BNN technology to perform traffic signal recognition tasks and propose the N+Half neural network model for the first time. The model only involves bit operations, addition and subtraction during inference. This design is extremely friendly to hardware resources and can significantly save chip area and reduce computing power consumption. Specifically, our traffic sign recognition model achieves significant storage compression ratios of 10.8x and 2.7x compared to the full-precision model and the int8 quantized model, respectively, and the accuracy loss is kept within 1% compared to the full-precision model.

## II. RELATED WORK

### A. *Model quantization and pruning*

Using convolutional neural networks for image classification is a widely used method, especially in the field of traffic sign recognition. The most commonly used dataset in this field is the GTSRB dataset proposed in [8]. Previously, the multi-column CNN network model proposed by [9] achieved an accuracy of 99.46% on the test set of the GTSRB dataset, which is an outstanding achievement in the history of convolutional neural networks. However, with the continuous advancement of research, [10] built a model using VisionTransFormer (ViT) technology, the model introduces the self-attention mechanism and performs different weighted calculations on the importance of each part of the input data, further improving the accuracy to 99.58%. The general research consensus is that the complexity of the network structure is often positively correlated with the accuracy of the model.

Quantized neural networks have become a research focus in recent years. Researchers have gradually realized that although complex models can achieve better accuracy, their deployment on embedded platforms often faces challenges [11]. Quantized models are mainly divided into two categories: post-training quantization and quantization-aware training [12]. Post-training quantization methods can leverage pre-trained models, but usually result in a reduction in model accuracy. In contrast, quantization-aware training requires additional training steps but often achieves higher accuracy. In order to effectively alleviate the problem of accuracy loss in the quantization process, [13] proposed STE, which can significantly improve the performance of the quantized model.

In the process of implementing the convolution operator, due to the issue of universality, most neural network hardware acceleration platforms will not directly implement the convolution operation, but will implement the more common matrix multiplication operation. The process from convolution operation to matrix multiplication operation generally uses Img2col [14] for conversion, which will convert the multi-channel input and convolution kernel into input matrix and convolution kernel matrix. Generally speaking, such a conversion will bring greater data depth because this conversion will store more data.

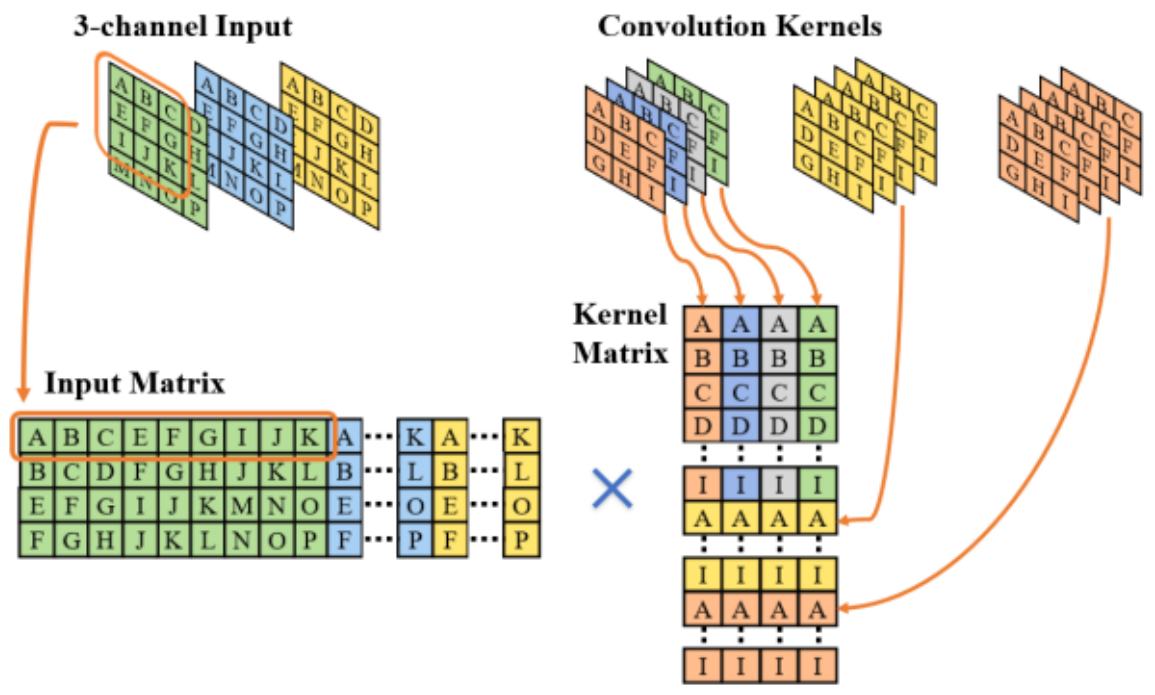

Fig. 1. GEMM-based convolution and img2col.

### B. *Binary Neural Network*

BNN is undoubtedly a groundbreaking work. Since BNN was first proposed in [5], it has attracted many researchers to optimize it from different angles. Among them, [15] proposed an innovative network model that only binarizes the feature maps, while the weight parameters are kept as fixed-point numbers with a higher bit width. This improvement effectively improves the accuracy of the model. In addition, [16] filled the gap in BNN research in the field of self-supervised learning, proposed S2-BNN, and demonstrated the performance of BNN on the ImageNet [21] dataset for the first time. On the other hand, the XOR-net proposed in [17] deeply considered quantization errors problem and enhanced the expressiveness of the model by introducing a scaling factor on each output channel, which enabled the BNN model to achieve remarkable results on the ImageNet dataset. The IR-net proposed in [18] adopted different optimization strategies in forward propagation and backpropagation: in forward propagation, Libra-PB was used to minimize the quantization loss of weights; in backpropagation, the designed EDE strategy aims to minimize the information loss of gradients. These two strategies are very simple, easy to understand and very effective.

In recent years, optimization technologies in related fields have emerged in an endless stream. ReActNet is a new network designed in [19]. Its core innovation is the introduction of the Bias term in PRelu and Sign functions. It is worth noting that most of the research works mentioned above include batch normalization layers. However, the study [20] attempted to remove the batch normalization layer from the network model because it may introduce additional floating-point operations, thus affecting the efficiency of training and inference.

The dataset GTSRB used in this paper is an image classification dataset covering 43 categories. In [22], researchers preprocessed the input image using image enhancement technology, and then used CNN for model training, and finally achieved an accuracy of 96%. [23] proposed using VIT algorithm to classify the dataset, which also achieved impressive accuracy performance. CTS is a Chinese traffic sign recognition dataset with 58 categories [24], and the BTS dataset is a Belgian traffic signal recognition dataset [25], which contains 62 categories of images. We use this dataset to verify the stability of our model. However, it is worth noting that most of the current research on the GTSRB dataset focuses on improving accuracy, while relatively few considerations are given to how to lightweight the model when deploying it on a vehicle platform. Although [26] proposed a signal recognition solution designed for AIOT devices, which has a relatively simple model structure and operation process, this scheme did not fully consider the additional overhead that may be caused by the data bit width dispersion during the reasoning process. The IE-Net mentioned in [27] proposed an information enhancement binary convolution to address the information loss caused by binarization of weights and activations, which can improve the performance of the model. Then, the information enhancement estimator IEE was introduced to approximate the sign function to reduce the amplitude of information attenuation, which has been proven to be effective on some common datasets.

## III. METHODOLOGY

This section describes our model structure and working principle in detail. We first proposed the N+Half network structure, which can eliminate the floating-point operations in the last layer of the inference stage. In addition, our design combines the operation fusion technology and distribution adjustment strategy to reduce the bit width of the values actually involved in the calculation in the inference phase, greatly reducing the storage pressure of the model inference stage.

### A. *N+Half Model*

The neural network structure used in this paper is called N+Half. Its overall structure is shown in Fig 2. The left side shows the structure of the model in the training phase, and the right side shows the structure of the model in the inference phase. In general, after the training phase is completed, the model can use the operation fusion method to obtain a more concise operation data flow.

N corresponds to N layers of convolution blocks. A convolution block includes: convolution layer, pooling layer, activation function layer, batch normalization layer. Since the fully connected layer, as the main content of the last part of most neural networks, will bring a lot of calculations, making the design of the computing component PE complicated, we canceled the fully connected layer structure in the design. Specifically, the first M layers use 2D convolution blocks, and the following K layers use 1D convolution blocks. The output dimension of the last layer of 1D convolution blocks matches the final number of categories. In this way, PE only needs to support convolution operations, which can make PE more concise.

Half refers to the incomplete calculation block. The incomplete calculation block only contains convolution layer and pooling layer. It is located after the last convolution block and is the last layer of the entire network. Since floating-point arithmetic operations consume several times more resources than fixed-point operations in FPGA or ASIC platforms, in order to avoid this situation, we introduce incomplete calculation block to completely eliminate the floating-point operations in the reasoning stage. The specific reasoning proof process will be introduced in Section 3 of this chapter.

The model file obtained after the training phase is completed will be used in the inference and test phases. We exported the model parameters and wrote a function to simulate the actual operation process of the hardware platform in the inference phase.

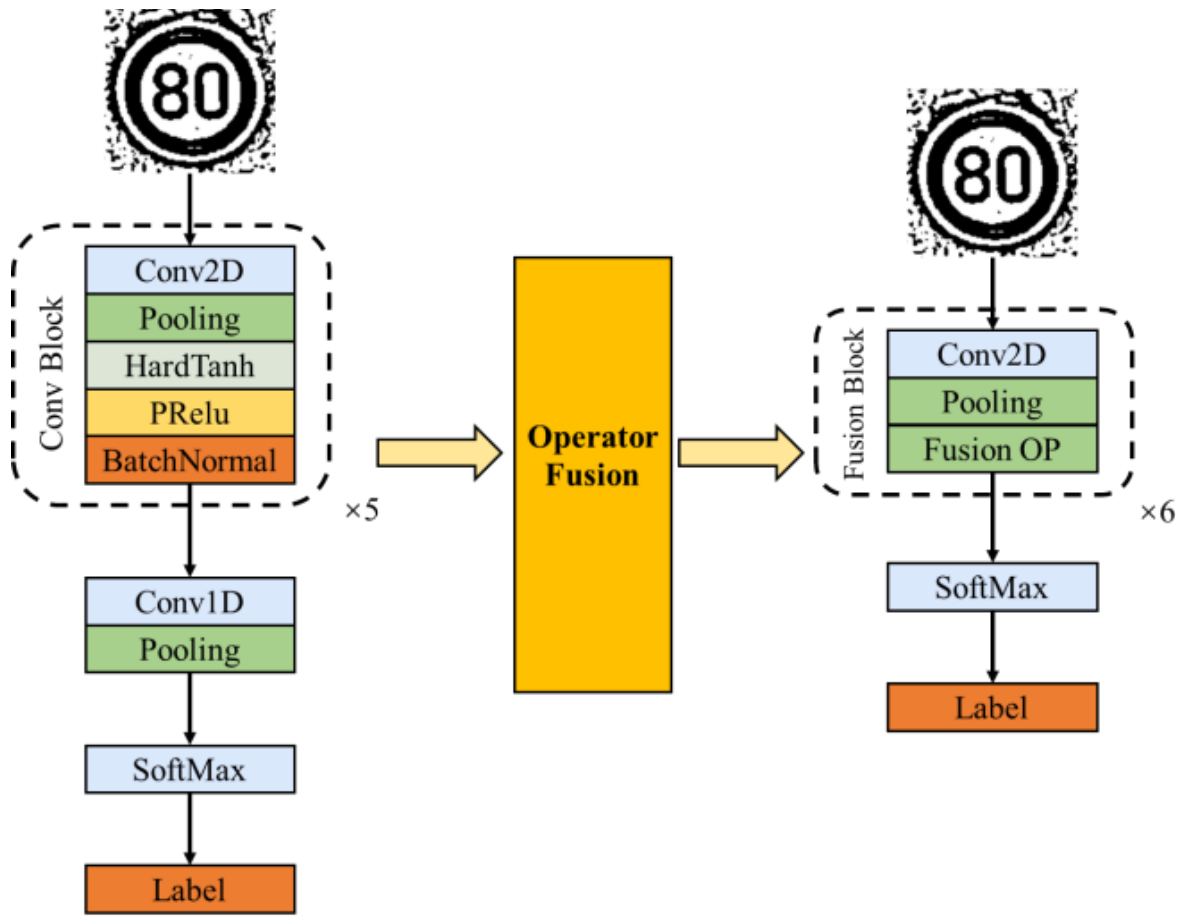

Fig. 2. N+Half BNN Model structure and fusion effect

### B. *Method of Binarization*

Traditional quantization methods can quantize the original floating-point parameters into int8 fixed-point numbers. This conversion brings many advantages, especially in FPGA or ASIC design, because floating-point Point operations consume a lot of resources. For BNN, its greatest contribution is to convert the multiplication and accumulation operations of convolution layer and fully connected layer in the original quantized network structure into logical operations such as XOR, thereby greatly reducing the computational complexity and resource occupancy. We use Formula 1 to convert both weights and inputs into {+1, -1} values, thereby binarizing the network.

$$x^b = Sign(x^r) = \begin{cases} +1, & if\, x \geq 0 \\ -1, & otherwise \end{cases} \qquad (1)$$

In Formula 1, $x^r$ is the original value and $x^b$ is the quantized value. Since the Sign function is not differentiable at 0, the chain rule does not work here. We use the IEE method proposed in [27] to approximate the derivative so that the neural network model can be back-propagated smoothly.

Fig 3 shows the function graphs of different activation functions. We first exclude the Tanh function because its operation is too complicated. The Sign function is used to obtain the binary parameters, and the HardTanh function is used to limit the bit width of the intermediate results. They play a key role in the model.

| Operation | Function plots | Derivative plots |
|---|---|---|
| $Tanh(x) = \frac{e^x - e^{-x}}{e^x + e^{-x}}$ | 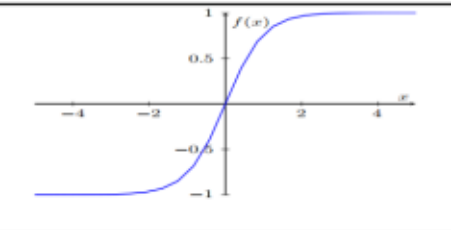 | 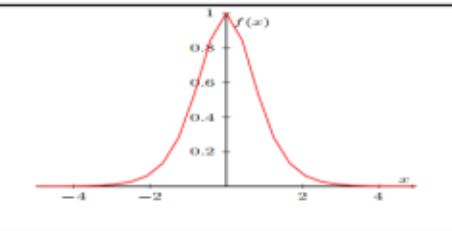 |
| $sign(x) = \begin{cases} +1 & x \geq 0 \\ -1 & x < 0 \end{cases}$ | 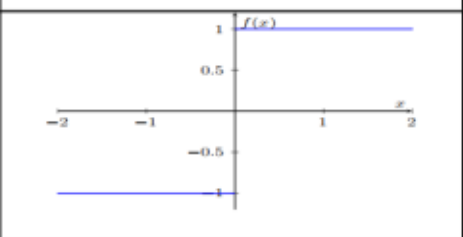 | 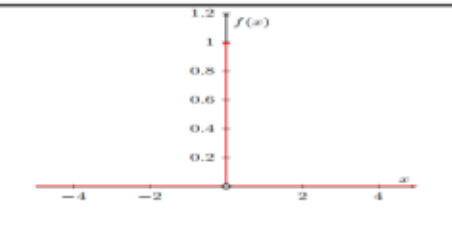 |
| $HTanh(x) = \begin{cases} +1 & x > 1 \\ x & -1 \leq x \leq 1 \\ -1 & x < -1 \end{cases}$ | 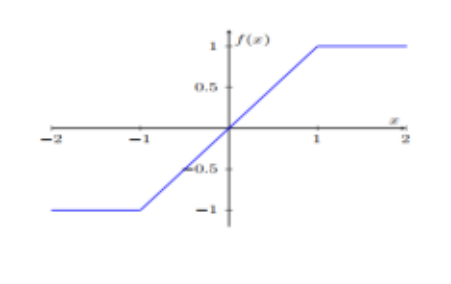 | 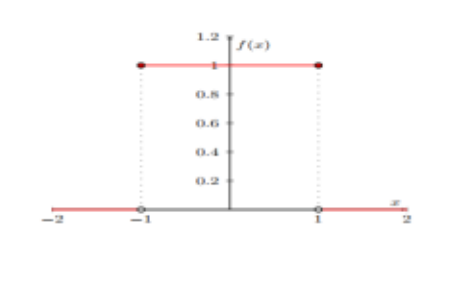 |

Fig. 3. Comparison between activation function Tanh, sign and HardTanh.

## C. Operator Fusion

Through operation fusion technology, we can integrate the operation process of batch normalization and activation function into only addition and subtraction operations. This strategy greatly simplifies the design of the operation component PE. In the network model, multiplication and division operations usually take up more resources when implemented on the hardware platform, and may reduce the operation speed, thereby restricting the operating frequency of the chip.

In order to improve model performance and accelerate convergence, we introduced two activation functions, Prelu and HardTanh, as well as a batch normalization layer to adjust the distribution of data. The core of operation fusion is to integrate these operations into one formula, thereby avoiding unnecessary multiplication and division operations and limiting the entire calculation process to addition and subtraction. Specifically, the process of the HardTanh function is shown in Formula 2.

$$HardTanh(x) = \begin{cases} maxval & if\ x > maxval \\ minval & if\ x < minval \\ x & otherwise \end{cases} \quad (2)$$

The HardTanh function means a clipping operation that limits the value to a specified range, which is generally symmetrical, such as -31 to 31. In our model, HardTanh is located after the pooling layer and before the Prelu activation function, which allows the range of parameters to be fixed within a certain range.

$$PReLU(x) = \begin{cases} x, & if x \geq 0 \\ ax, & otherwise \end{cases} \quad (3)$$

$$BatchNormal(x) = \frac{x-\mu}{\sqrt{\sigma^2+\epsilon}} * \gamma + \beta \quad (4)$$

Formula 3 is the mathematical formula of the Prelu function, where a is a trainable parameters whose optimal value can be found during the training process. Formula 4 is the mathematical formula of the batch normalization layer, where $\mu$ and $\sigma$ are the mean and variance of the data. Since the training data and test data generally have similar distributions, they are also trainable parameters. $\varepsilon$ is a constant that is usually used to avoid division by 0. In addition, $\gamma$ and $\beta$ are also trainable parameters. If the network is built in the order of HardTanh, Prelu, and batch normalization, the complete mathematical formula can be considered as shown in Formula 5.

$$Func(x) = \begin{cases} C1 & if x > maxval \\ C2 & if x < minval \\ kx + b & if x > 0 \\ akx + b & if x < 0 \end{cases} \quad (5)$$

$$k = \frac{\gamma}{\sqrt{\sigma^2+\varepsilon}} \quad (6)$$

$$b = \beta - \frac{\mu\gamma}{\sqrt{\sigma^2+\varepsilon}} \quad (7)$$

$$C1 = maxvalue * k + b \quad (8)$$

$$C2 = a * minvalue * k + b \quad (9)$$

Formula 5 means that if the content of *x* exceeds the upper and lower limits set previously, it will be set to constant C1 or C2. In other cases, depending on the sign bit of x, two sets of operations will be involved, in which the parameters k and b involved in the operation are composed of trainable parameters. In the inference phase, k and b are also constants. After the above operation fusion, the division operation in the batch normalization of the inference phase can be eliminated, but because of the existence of a, floating-point multiplication still exists. However, due to the existence of the subsequent Sign function, this problem will also be solved. Here we need to introduce a new function Sign-Plus, as shown in the process Formula 10 and Formula 11.

$$Sign^{+}(x, \delta) = \begin{cases} +1, & if x \geq \delta \\ -1, & otherwise \end{cases} \quad (10)$$

$$Sign^{-}(x, \delta) = \begin{cases} +1, & if x \leq \delta \\ -1, & otherwise \end{cases} \quad (11)$$

Combining the content of Formula 10, Formula 11, Formula 5 and Formula 1 are combined to obtain the following Formula 12.

$$Final(x) = \begin{cases} Sign(C1) & if x > max_v al \\ Sign(C2) & if x < min_v al \\ Sign^{+}(x, \delta 1) & if x > 0 and k > 0 \\ Sign^{-}(x, \delta 1) & if x > 0 and k < 0 \\ Sign^{+}(x, \delta 2) & if x > 0 and k > 0 \\ Sign^{-}(x, \delta 2) & if x > 0 and k < 0 \end{cases} \quad (12)$$

$$\delta 1 = \frac{-b}{k} \quad (13)$$

$$\delta 2 = \frac{-b}{ak} \quad (14)$$

Combined with Formula 12, it can be concluded that the operation after the pooling operation in the inference phase only needs to compare the *x* and some constants and the sign bit of k to complete the operation process. In our neural network model, the Sign function is placed in the next Block convolution layer, which is why the N+Half structure is designed, because only in this way can the last operation fusion be completed normally. If there is no last incomplete block, the last round of fusion will lack a Sign function, so only Formula 5 can be used for calculation instead of Formula 12, but Formula 5 essentially contains floating-point multiplication, which is very disadvantageous in hardware deployment. In summary, the structure of the convolution Block should be: convolution layer, pooling layer, HardTanh, Prelu and batch normalization.

## D. Adjusted Distribution

In the convolution operations, although the weights and input data are both 1 bit wide, their results are usually not. These results need to be temporarily stored for subsequent operations and quantization. By adjusting the distribution strategy, we can reduce the storage space of these intermediate results to the ideal range, thereby saving a lot of storage space.

The following figures (Fig.4, Fig.5 and Fig.6) show the comparison of the numerical distribution of *x* and two parameters used for comparison when the HardTanh layer is introduced. In Fig 4, the right side shows the situation without HardTanh being introduced, and the left side shows the situation after the introduction. It can be observed that the introduction of HardTanh can significantly limit the distribution range of the pooling layer output data, so that these data can be stored with a smaller bit width, thereby significantly easing storage pressure. In Fig 5, it can be observed that the distribution of data is concentrated in the specified range, because the distribution of parameters depends on the threshold set. When the threshold value is 31, 6bit is needed to store parameters. In Fig 6, it can be observed that the distribution of data is very loose. In this case, 15bit is needed to store parameters.

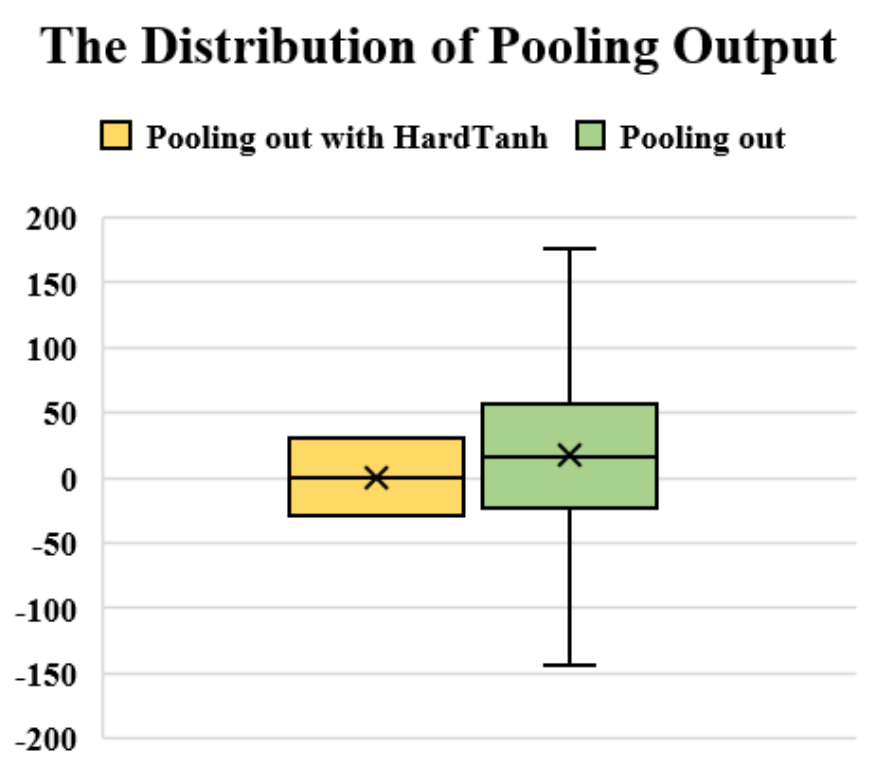

Fig. 4. The distribution of the pooling layer output data.

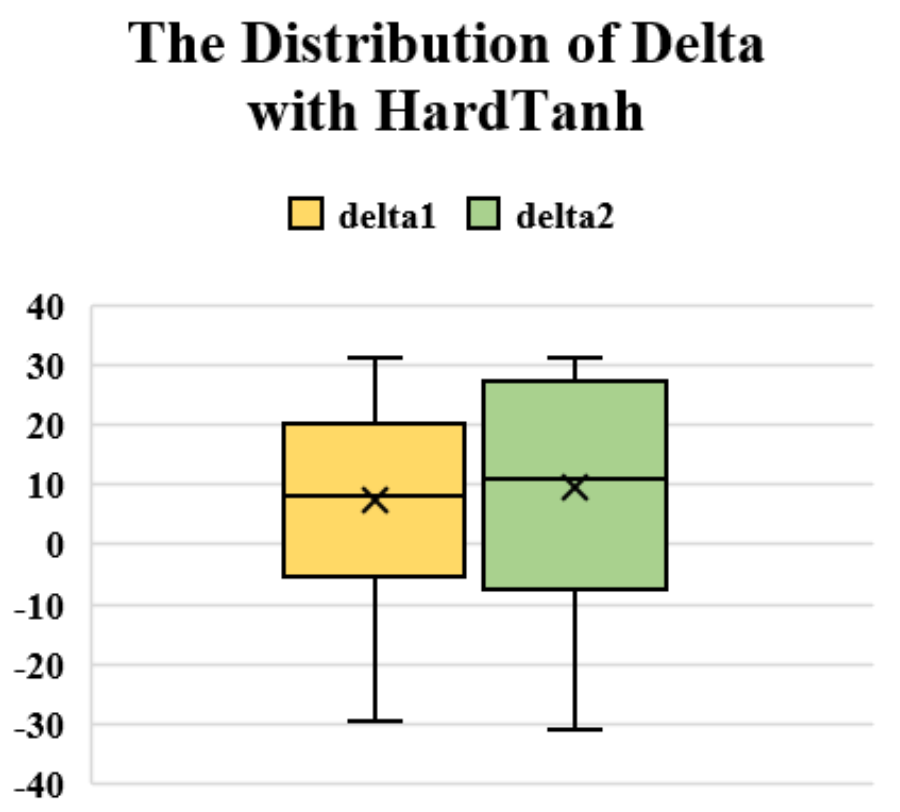

Fig. 5. The distribution of parameters when HardTanh is introduced.

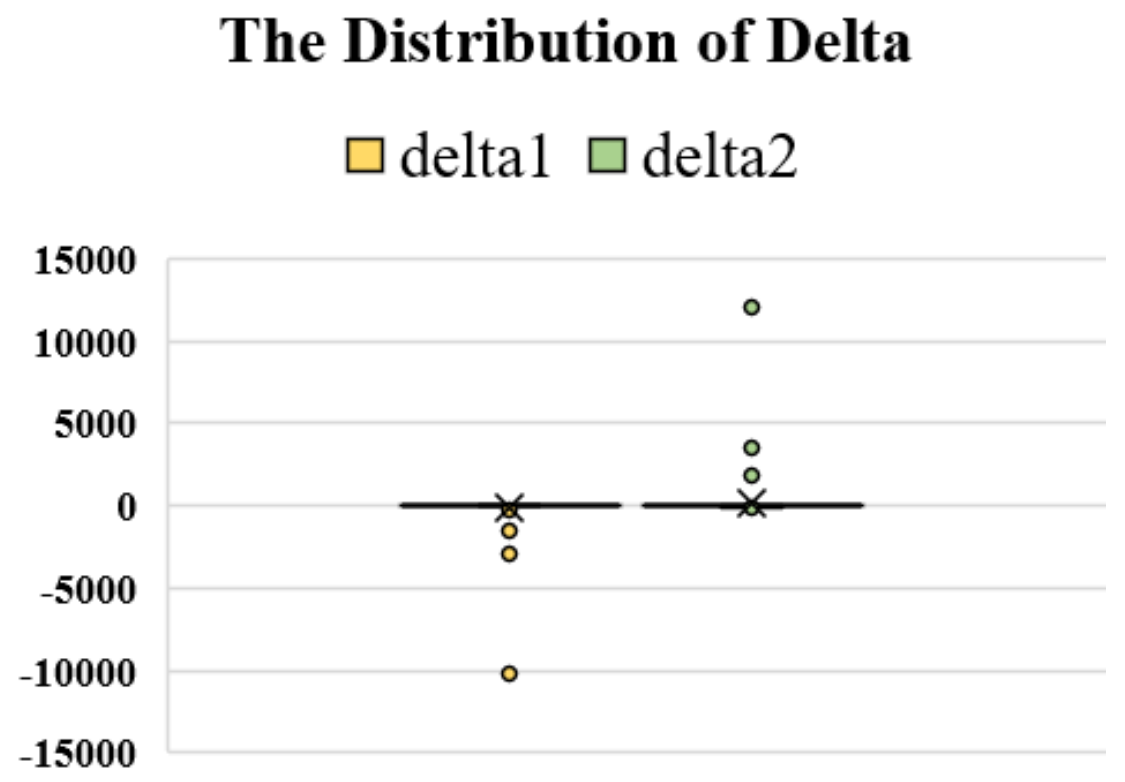

Fig. 6. The distribution of parameters without the introduction of HardTanh.

## IV. EXPERIMENT

In this section, we will first introduce the dataset used in this paper and the setting of model parameters during the experiment. Then, we will introduce the model test results and the comparison results with other models. Finally, we will introduce the ablation experiment to demonstrate the irreplaceability of each module in the model and the rationality of the hyperparameter values in the experiment.

### A. DataSet and preprocess

The neural network model used in the experiment works on the GTSRB dataset, which is a traffic signal recognition dataset with 43 categories. The entire dataset has more than 50,000 images for training and testing, and most of the image data is captured by on-board cameras, which is closer to the actual situation of autonomous driving. The training and reasoning process of this paper will use OpenCV to preprocess the image and obtain a grayscale image of the specified size. After obtaining the grayscale image, the pixel matrix will be divided by 255 as a whole to obtain a binary pixel matrix.

Fig 7 intuitively shows the effect of this preprocessing. Because a binary neural network is used, after the original image is binarized, the 0 in the pixel value will be replaced by -1. In this way, both the model weights and the input image are numerically unified, containing only +1 and -1.

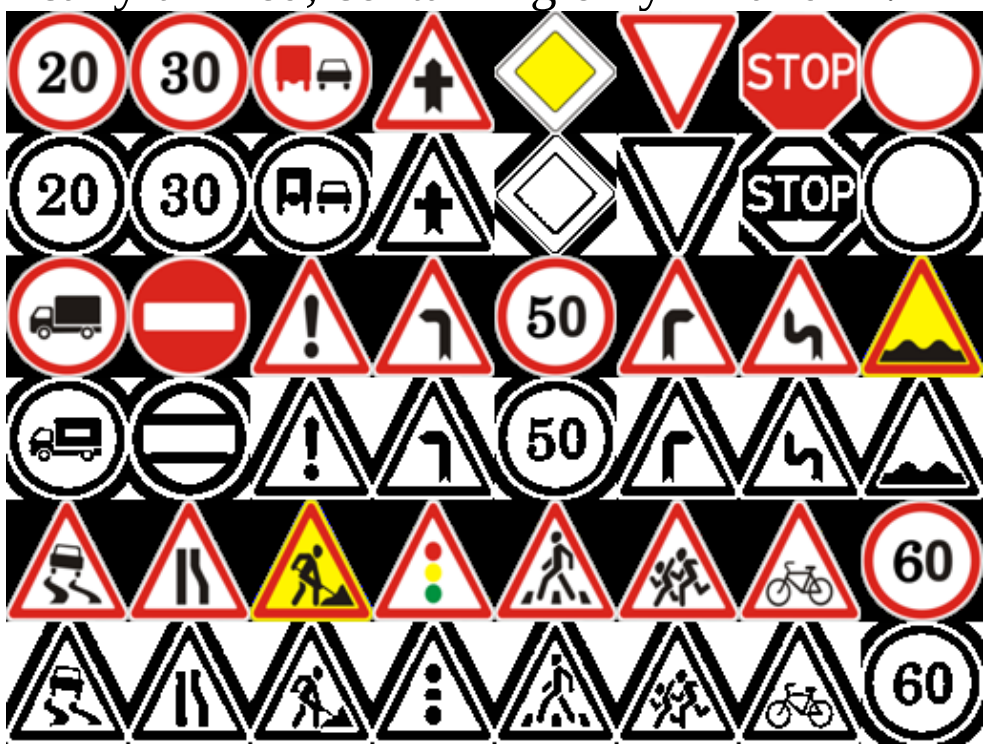

Fig. 7. Comparison of original images and preprocessing results on the GTSRB dataset.

### B. Model structure and parameters

After multiple experimental verifications, we determined that the optimal number of blocks for the BNN network structure is 6. This choice is based on the N+Half structure we adopted, and it is worth noting that the 6th block is designed as an incomplete calculation block. Table 1 lists in detail the specific parameter configurations of the convolutional layer in each block in the network structure. Based on this optimal structure, we can achieve an accuracy of 97.19% in the training phase. Combined with the operation fusion method mentioned above, an accuracy of 96.76% can be achieved in the inference phase on the test set. In comparison, on Kaggle, the SOTA method achieves an accuracy of 98.06% on the CNN full-precision model of the GTSRB dataset. The same network structure was used to test the CTS and BTS datasets, and the accuracy rates were 98.58% and 96.89% respectively during the training phase.

TABLE I. BNN MODEL PARAMETERS

| Table Head | In Channel | Out Channel | Conv Kernel Size | Conv Stride | Pool Size | Pool Stride |
|---|---|---|---|---|---|---|
| Block1 | 1 | 8 | 5 | 1 | 2 | 2 |
| Block2 | 8 | 16 | 5 | 1 | 2 | 2 |
| Block3 | 16 | 32 | 5 | 1 | 2 | 1 |
| Block4 | 32 | 64 | 5 | 1 | 2 | 1 |
| Block5 | 64 | 128 | 16 | 1 | 4 | 2 |
| Block6 | 128 | 43 | 16 | 1 | 4 | 2 |

### C. Performance Comparison

In terms of the storage space occupied by the convolutional layer parameters, the binary parameters of the BNN network are 287032, which will occupy 10KB of space. Compared with the full-precision network model, the space compression ratio is 10.8. Compared with the quantized model obtained by int8 quantization of the full-precision network model, the space compression ratio is 2.7. In the additional operations brought by the activation function after operation fusion, the spatial compression ratio for the storage of the intermediate results and comparison parameters of the feature map is 2.5. This is mainly due to the smaller bit width can be used to store intermediate results and comparison parameters.

Table 2 shows the comparison results of the proposed model with other lightweight models and full-precision models. For fairness, the initialization scale of the image is set to 128*128. In terms of accuracy, our accuracy is in the leading position in the BNN network. In terms of implementation, each BNN model of the convolution layer is implemented using XOR. In the subsequent batch normalization and activation function implementation, due to the operation fusion technology of this paper, our model can only use addition operations to implement the remaining operations, which can bring great advantages in the process of hardware implementation.

TABLE II. COMPARISON WITH OTHER MODELS

| **Work** | **Features** | **BitWidth(W/A)** | **ACC** | **Params/million** | **Params/KB** |
|---|---|---|---|---|---|
| Luca | robust BNN | 1bit/1bit | 92% | 4.3 | 524.9 |
| Andreea | Qconv | 1bit/1bit | 96.45% | 1.77 | 206.06 |
| Ours | N+Half | 1bit/1bit | **97.64%** | **0.287** | **35.03** |
| Sharma | CNN | 32bit/32bit | 98.06% | 0.097 | 380.625 |
| Shravan | ASG-ViT | 32bit/32bit | 99.58% | 8.6 | 33593.75 |

TABLE III. COMPARISON WITH OTHER MODELS

| **IP** | **RH** | **IB** | **Train_ACC** | **Test_ACC** | **Float_OPS** | **BitWidth** |
|---|---|---|---|---|---|---|
| YES | NO | NO | 98.53% | 98.22% | 22016 | 15 |
| YES | YES | NO | 97.81% | 97.23% | 22016 | 6 |
| YES | YES | YES | 97.64% | 96.16% | 0 | 6 |

### *D. Performance Comparison*

In order to evaluate the effectiveness of each component of the proposed neural network model, we decomposed the algorithm and hid the functions of the components one by one to verify the effectiveness of each component. The evaluation table criteria are mainly the train accuracy in the training phase, the test accuracy in the test phase, the number of floating-point operations Float_OPS in the model inference phase, and the bit widths of the intermediate result storage. Table3 shows the comparison results. We use YES to indicate the use of the module function and NO to indicate the hiding of the module function. The table shows that the introduction of RH and IB will affect the accuracy of the model, but the performance of the other two indicators will bring greater benefits. In order to facilitate hardware deployment, these accuracy losses are acceptable. Considering all indicators comprehensively, the best performance can be achieved by using all modules. These experimental results show that the image preprocessing module IP, the threshold adjustment structure RH and the incomplete operation block IB have made important contributions to the performance of the algorithm and played an irreplaceable role.

In order to verify the impact of the threshold value of the HardTanh threshold function on the accuracy of the model, we conducted an ablation experiment on the threshold for GTSRB, CTS and BTS. From the results in the Figure 8, it can be seen that when the absolute value of the threshold exceeds 31, the increase in the threshold range has little effect on improving the accuracy, and using a larger threshold does not achieve better model performance. Similarly, it can be observed that selecting a threshold that is too small will also limit the performance of the model. For example, when the threshold is set to 8, the model basically loses its predictive ability. Therefore, the absolute value of the threshold of the threshold function we finally chose is 31, which will limit the range of the intermediate results of the model calculation to {-31, 31}. This choice is both reasonable and effective.

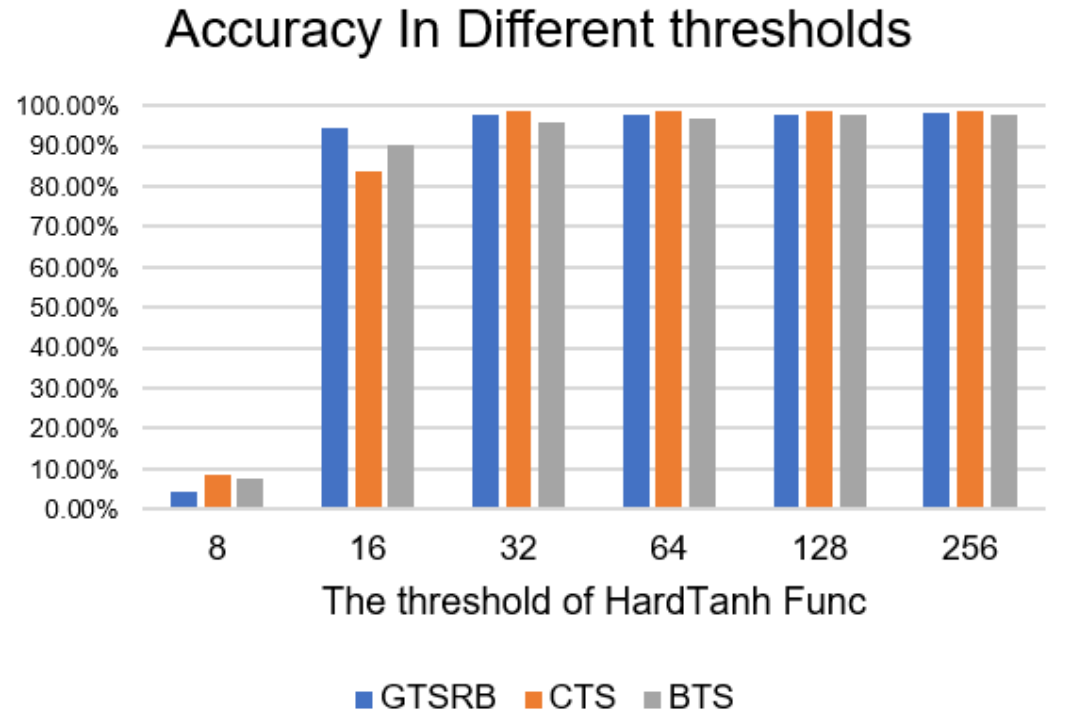

Fig. 8. The relationship between the accuracy rate and threshold.

## V. CONCLUSION

In this paper, we successfully achieved higher recognition accuracy by binarizing feature maps and weights on the GTSRB traffic signal recognition dataset and cleverly integrated the subsequent activation function and batch normalization layer, and achieved the best performance in the binary neural network (BNN) task on this dataset. Specifically, we innovatively proposed the N+Half neural network model, which, in addition to convolution and pooling operations, only requires addition and subtraction operations to complete the calculation in the inference stage, greatly simplifying the calculation process. In addition, we also effectively reduced the data storage cost by optimizing the distribution of intermediate results, which is of great significance for hardware deployment. Given that the weights and activation functions in the network are restricted to a single binary number, binary networks have shown great application potential on hardware platforms such as FPGA or ASIC. As a next step, we plan to verify the acceleration effect and energy-saving performance of BNN during inference on FPGA and ASIC.